\pdfoutput=1
\documentclass[runningheads]{llncs}
\usepackage{graphicx}
\usepackage{comment}
\usepackage{amsmath,amssymb} % define this before the line numbering.
\usepackage{color}
\usepackage{soul}
\usepackage{wrapfig}
\usepackage{subfig}
\usepackage{hyperref}

\begin{document}
\pagestyle{headings}

\title{Group Activity Prediction with Sequential Relational Anticipation Model}

\begin{comment}
\titlerunning{ECCV-20 submission ID \ECCVSubNumber} 
\authorrunning{ECCV-20 submission ID \ECCVSubNumber} 
\author{Anonymous ECCV submission}
\institute{Paper ID \ECCVSubNumber}
\end{comment}
%******************

% \begin{comment}
\titlerunning{Group Activity Prediction with Sequential Relational Anticipation Model}
\author{Junwen Chen\orcidID{0000-0001-9808-1520},
Wentao Bao\orcidID{0000-0003-2571-3341},\\ \and
Yu Kong\orcidID{0000-0001-6271-4082}}

\authorrunning{J. Chen et al.}
\institute{Golisano College of Computing and Information Sciences\\ Rochester Institute of Technology\\
% \url{http://www.springer.com/gp/computer-science/lncs} \\
\email{\{jc1088,wb6219,yu.kong\}@rit.edu}}
% \institute{Rochester Institute of Technology, Princeton NJ 08544, USA \and
% \end{comment}
%******************
\maketitle

\begin{abstract}
In this paper, we propose a novel approach to predict group activities given the beginning frames with incomplete activity executions. 
Existing action\footnote{\scriptsize{We define action as the behavior performed by a single person, and define activity as the behavior performed by a group of people.}} prediction approaches learn to enhance the representation power of the partial observation\footnote{\scriptsize{We define \emph{partial observation} as the beginning frames with incomplete activity execution, and \emph{full observation} as the one with complete activity execution.}}.
However, for group activity prediction, the relation evolution of people's activity and their positions over time is an important cue for predicting group activity. To this end, we propose a sequential relational anticipation model (SRAM) that summarizes the relational dynamics in the partial observation and progressively anticipates the group representations with rich discriminative information.
Our model explicitly anticipates both activity features and positions by two graph auto-encoders, aiming to learn a discriminative group representation for group activity prediction.
Experimental results on two popularly used datasets demonstrate that our approach significantly outperforms the state-of-the-art activity prediction methods. \\
\end{abstract}

\section{Introduction}

Group activity prediction is to forecast an activity performed by a group of people before the activity ends. Different from group activity recognition,
it only has access to the beginning frames of a video containing \textbf{incomplete} activity execution. 
It is useful in the scenarios where the intelligent systems have to make prompt decisions, such as surveillance and traffic accident avoidance where multiple people are present.
Unfortunately, existing action prediction methods~\cite{yao2018multiple,yan2017predicting,kong2017deep,kong2014discriminative,kong2018action} are limited to actions performed by an individual. 
Even though some methods \cite{kong2017deep,kong2018pami,wang2019progressive} attempt to predict actions performed by multiple people in standard databases such as UCF101~\cite{soomro2012ucf101}, 
they simply model the multiple people as a single entity and ignore their relations. This would undoubtedly result in a low prediction performance.

\begin{figure}[!t]
\begin{center}
\includegraphics[width=0.85\linewidth]{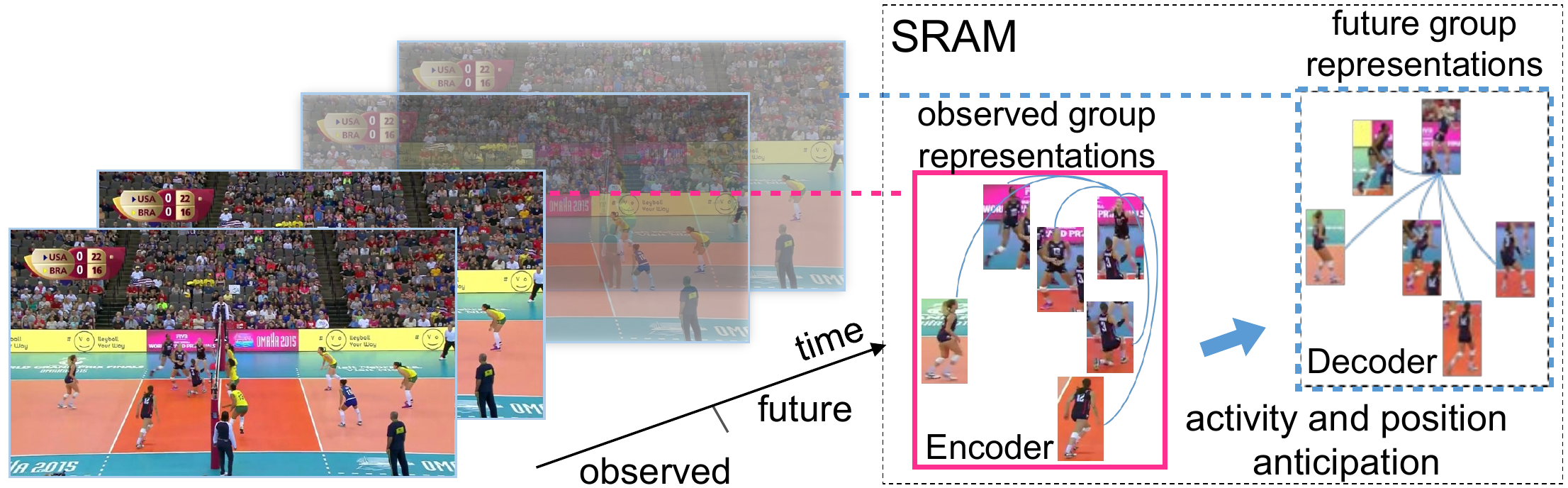}
\end{center}
\caption{
Given the beginning frames, our method models the relational dynamics of a group, and predicts a group activity by anticipating the group activity representation and their positions occurring in the future unobserved frames.
}
\label{fig:task1}
\end{figure}

As shown in~\cite{kong2017deep}, one of the major challenges in activity prediction is how to enhance the discriminative power of the features extracted from the partial observations. 
However, this is even more challenging to do so in group activity prediction as multiple people are present in the scene. Each person's individual action may vary and people's interactions frequently appear and change in a group activity. 
To this end, it is important to model the relations of multiple people in the observed frames and predict their future group representations. 
In addition, if only limited beginning frames are observed, it would be extremely difficult to directly anticipate the features of full observations at once. 
A temporally progressive anticipation model is desired for modeling activity evolution.

To address these challenges, we propose a novel sequential relational anticipation model (SRAM) for group activity prediction by anticipating group activities and positions in the future (see Fig.~\ref{fig:task1}). 
SRAM is developed as an encoder-decoder framework, in which an \textit{observation encoder} summarizes the relational dynamics in the beginning observed frames and a \textit{sequential decoder} further anticipates the representations for group activities and positions occurring in the future.
Specifically, the observation encoder naturally models the relational dynamics of people and complex interactions between people in the observed frames. 
To predict group activity, we propose a sequential decoder to anticipate the structured group representation in the future using several unrolling stages. Two graph auto-encoders are used in the sequential decoder to explicitly anticipate the activity and the position relations of people in the unobserved frames.
We propose to make sequential prediction that \emph{progressively} anticipates the future group representation by performing multiple unrolling stages guided by three novel loss functions. This allows us to better capture complex group activity evolution.

To our best knowledge, we are the first to investigate the challenging problem of group activity prediction. The benefit of our method is twofold. 
Firstly, it not only predicts people's group activities but also predicts individuals' positions in the future. Our experimental results show that predicting people's future positions significantly helps predict their group activities. 
Secondly, the proposed method progressively anticipates structured group representations, which has shown to be very powerful in prediction especially when limited frames are observed. This idea could be generalized to other prediction tasks, e.g. human motion prediction~\cite{martinez2017human} and video prediction~\cite{wichers2018hierarchical}.

Our contributions can be summarized as follows:
\begin{itemize}
    \item We propose a novel sequential decoder to anticipate the representations for multiple people's future positions and activity, aiming to learn a discriminative structural representation for group activity prediction.    
    \item We progressively anticipate the structured group representations at several unrolling stages guided by novel loss functions. This improves the performance when only few frames are observed. 
    \item Extensive experiments demonstrate that our method outperforms the existing state-of-the-arts by a large margin.
\end{itemize}

\section{Related work}
~\textbf{Action Prediction} aims to recognize the label of an action before the action is fully executed.
Existing work~\cite{lan2014hierarchical,cai2019action,kong2018action,hu2018early,sadegh2017encouraging,ma2016learning,zhao2019spatiotemporal,shi2018action,vondrick2016anticipating} focuses on predicting actions performed by an individual. 
Ryoo et al.~\cite{ryoo2011human} used integral and dynamic bag-of-words to represent features variations over time. 
DeepSCN~\cite{kong2017deep} and AAPNet~\cite{kong2018pami} make use of sequential context information by transferring knowledge in full videos to partial observations. 
Wang et al.~\cite{wang2019progressive} developed a teacher-student learning framework to distill knowledge from the action recognition task, in order to enhance action prediction. Gammulle et al.~\cite{gammulle2019predicting} presented a jointly learnt task for both action prediction and future motion representation inference.

Prediction on interactions between two people was studied in \cite{yan2017predicting,yao2018multiple}. Yan et al.~\cite{yan2017predicting} developed a tri-coupled recurrent structure and an attention mechanism to address action prediction for two individuals' interactions.
Yao et al.~\cite{yao2018multiple} predicted the motion of the interactions between two people, but did not predict their interaction labels. 
Different from them, we focus on the prediction of group activities involving multiple people. Our method elegantly captures complex relational dynamics between people for learning discriminative information.

~\textbf{Group Activity Recognition} has been extensively studied in previous work \cite{amer2014hirf,shu2017cern,ibrahim2016hierarchical,yan2018participation}.
Early work applies graphical models on the extracted hand-craft features~\cite{amer2014hirf,lan2012social,wang2013bilinear} as group representations. 
Deep learning methods for multi-people activity recognition have shown excellent performance~\cite{shu2017cern,bagautdinov2017social,wang2017recurrent,ibrahim2018hierarchical,deng2016structure,tang2018mining,Gavrilyuk_2020_CVPR}.
HDTM~\cite{ibrahim2016hierarchical} develops a two-stage LSTM model to firstly extract features of temporal individual motions and then aggregate neighborhood information.
SSU~\cite{bagautdinov2017social} achieves the individual detection and group activity recognition in a unified framework.
Recent work suggests that only part of people's motions contribute to the entire group activity~\cite{yan2018participation,gammulle2018multi,ramanathan2016detecting,azar2019convolutional,hu2019progressive}, via suppressing the irrelevant actions.
Previous work also shows that interactions between people are important in understanding group activity. 
For example, HRN\cite{ibrahim2018hierarchical} introduces a hierarchical spatial relational layer to learn the relational representations between two players. 
Other methods, including Stagnet~\cite{qi2018stagnet}, S-RNN~\cite{biswas2018structural}, SBGAR~\cite{li2017sbgar} apply structural-RNN to obtain spatiotemporal features. 
ARG~\cite{wu2019learning} explicitly models the interactions by employing graph convolution on a learnable graph.

The main difference between our work and group activity recognition methods is that we aim at predicting the group activity label given \emph{incomplete} activity execution, while these methods are given complete activity executions. This prompts us to develop novel model architecture and loss functions in this work.

\section{Our approach}

\textbf{Problem formulation.}
Our goal is to predict the activity label $y$ of a group of people given a partial observation of a video containing incomplete activity execution.
We define the \textit{observation ratio} as the number of observed frames $t_0$ in a streaming video divided by the total number of frames $T$ in the corresponding full video following \cite{kong2017deep,kong2018pami}, i.e. $t_{0} / T$. For instance, if a partial video contains $30$ frames and the corresponding full video contains $100$ frames, then the observation ratio of this activity is $30\%$. 

During training, we have access to all full training videos containing complete group activity executions. These full videos are supposed to contain all the discriminative information for classification.
During test, given a partial observation of a group activity execution, we encourage our model to anticipate the group representations that contain similar amount of discriminative information as the corresponding full observation. Thus, its prediction power can be enhanced.

\begin{figure*}[!t]
\begin{center}
 \includegraphics[width=1\linewidth]{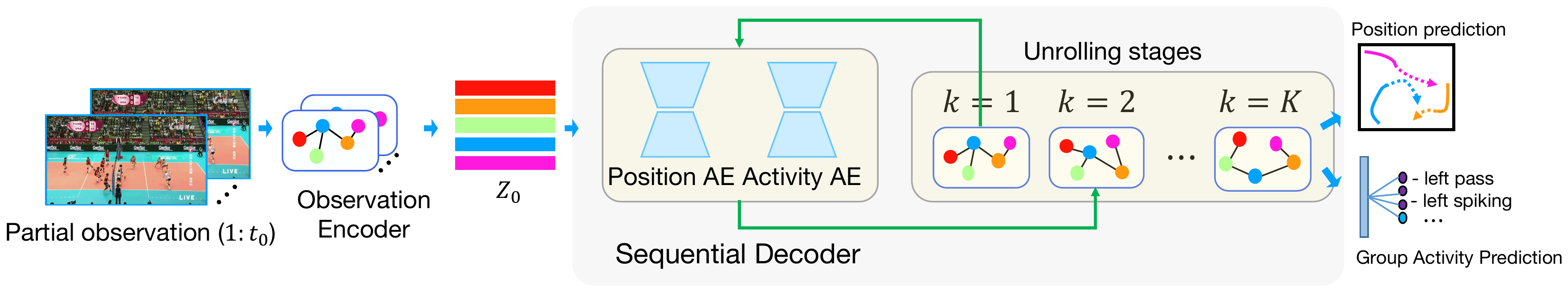}
\end{center}
\caption{
Overall architecture.
Our framework SRAM takes the beginning $t_0$ observed frames as input and predicts the group activity label. 
An observation encoder first summarizes the relational dynamics in partial observation as a latent variable $Z_0$. Then, a sequential decoder takes over $Z_0$  and progressively anticipates the group representation through $K$ unrolling stages. The output of the last unrolling stage is expected to contain rich discriminative information for group activity prediction.
Details can be seen in Fig.~\ref{fig:architecture3}.
}
\label{fig:approach2}
\end{figure*}

\textbf{Overall architecture.}
The overall architecture is shown in Fig.~\ref{fig:approach2}. We formulate our group activity prediction model as an encoder-decoder framework that contains an \emph{observation encoder} and a \emph{sequential decoder}. Given a partial observation containing $t_0$ frames, the observation encoder summarizes the relational dynamics of the group from the partial observation and then the sequential decoder anticipates the group representation for activities and positions in the future unobserved frames. 

Due to the large motion variations between a partial and a full observation, a novel sequential decoder is proposed in this work to progressively anticipate the structured group representation for the future unobserved frames by several unrolling stages. This is useful for enhancing the discriminative power of the anticipated representation, especially if limited frames are observed. 
Moreover, for group activity, relations between multiple people are discriminative information and they vary as time. To predict group activity, our sequential decoder uses two graph auto-encoders to concurrently perform relational anticipation on both people's activity features and their positions.

\subsection{Relation Modeling for Group Activity}

Given $t_0$ observed frames, we first extract features of all the observed $t_0$ frames, and then apply ROIAlign~\cite{he2017mask} to extract the feature vectors of multiple people based on their positions $\left \{B_1,B_2,\cdots,B_{t_0}\right \}$($t\in\{1,\cdots,t_0\}$). Action features and positions of the $i$-th individual on the $t$-th frame are represented as $\mathbf{x}_t(i)$ and $\mathbf{b}_t(i)$ respectively. 
Afterwards, upon the individual dynamics, we follow~\cite{wu2019learning} to explicitly model the pair-wise \textit{position relations} and \textit{action relations} of multiple people in the observed frames as two relation graphs $G^{\text{a}}_{t}\in \mathbb{R}^{N\times N}$ and $G^{\text{p}}_{t}\in \mathbb{R}^{N\times N}$ , respectively. Both of the two graphs have $N$ nodes representing $N$ people in the $t$-th frame.
Given the $i$-th and $j$-th individuals, the edge of the action similarity graph $G^{\text{a}}_{t}(i,j)$ is computed by the cosine similarity and normalized by Softmax function. The edge on the position relation graph $G^{\text{p}}_{t}(i,j)$ is computed by the normalized Euclidean distance (denoted by $d(\cdot, \cdot)$):
\begin{equation}
\label{eq:normalizeAPP}
G^{\text{a}}_{t}(i,j)=\frac{\exp\left ( \mathbf{x}_{t}(i)^{\textrm{T}}\cdot \mathbf{x}_{t}(j) \right )}{\sum_{j=1}^{N}\exp\left ( \mathbf{x}_{t}(i)^{\textrm{T}}\cdot \mathbf{x}_{t}(j) \right )}, \:
G^{\text{p}}_{t}(i,j) = \frac{1/d(\mathbf{b}_{t}(i),\mathbf{b}_{t}(j))}{\sum_{j=1}^{N}1/d(\mathbf{b}_{t}(i),\mathbf{b}_{t}(j))}.
\end{equation}

Once the graphs are built, we obtain the structured representations for the group activity in the observed frames. We will also perform anticipation on the two graphs representing the group activity in the unobserved frames, which will be discussed below.

\subsection{Observation Encoder $\mathcal{E}$}
The observation encoder $\mathcal{E}$ is proposed to summarize spatiotemporal information of the complex relational dynamics of multiple people in partial observations containing $t_0$ frames. 
$\mathcal{E}$ learns to map $G^{\text{a}}_{1:t_0}$, $G^{\text{p}}_{1:t_0}$, and $X_{1:t_0}$ to a latent variable $Z_0$, by the spatio-temporal graph convolution network ST-GCN~\cite{yan2018spatial}. Specifically, it first performs 
spatial graph convolution~\cite{kipf2016semi} on the two graphs $G_{t}^{\text{p}}$ and $G_{t}^{\text{a}}$ for the $t$-th frame
\begin{equation}
\label{eq:oe}
\sigma (G_{t}^{\text{p}}X_{t}W_\text{p})+\sigma(G_{t}^{\text{a}}X_{t}W_\text{a}),
\end{equation}
and then performing temporal convolution \cite{lea2017temporal} on every three consecutive frames to learn the latent variable $Z_0$. Here, $\sigma$ is ReLU activation, $W_\text{p}$ and $W_\text{a}$ are learnable weights, and $X_t$ is the action features of $N$ people. Latent variable $Z_0$ will be integrated in the sequential decoder, and guides its unrolling stages.

Different from ARG~\cite{wu2019learning}, our model captures the temporal patterns of people's relations, which is useful for group activity prediction.

\subsection{Sequential Decoder $\mathcal{D}$}
\begin{figure}[!t]
\begin{center}
\includegraphics[width=0.8\linewidth]{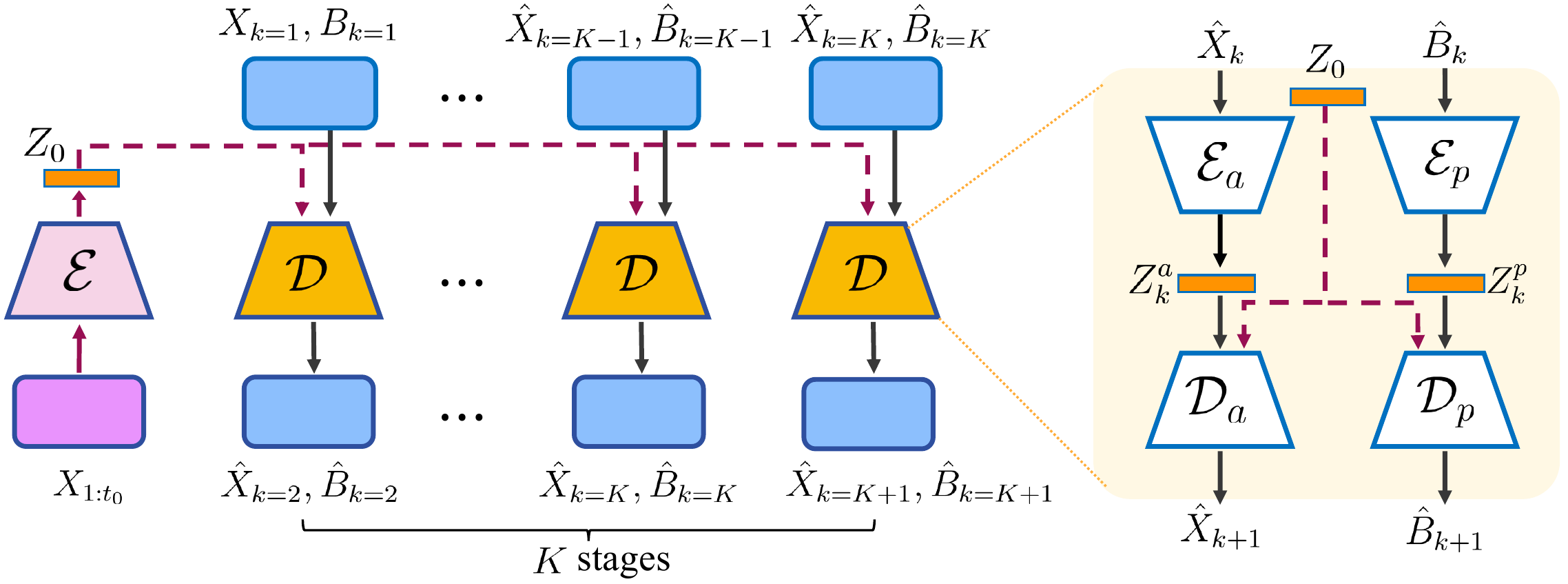}
\end{center}
\caption{
Sequential decoder $\mathcal{D}$ is formulated as two auto-encoders $\mathcal{E}_\text{a}\text{-}\mathcal{D}_\text{a}$ and $\mathcal{E}_\text{p}\text{-}\mathcal{D}_\text{p}$ that progressively anticipate the group activity representation for future unobserved frames using multiple unrolling stages. 
At the $k$-th stage, $\mathcal{D}$ is fed with the summary of the partial observation encoded in the latent variable $Z_0$ as well as the action features $\hat{X}_{k}$ and the position features $\hat{B}_{k}$ from the previous stage. Then $\mathcal{D}$ anticipates the action features $\hat{X}_{k+1}$ and positions $\hat{B}_{k+1}$.}
\label{fig:architecture3}
\end{figure}

The performance of state-of-the-art action prediction methods \cite{kong2018pami,kong2017deep} is still limited especially when few beginning frames are given. This is mainly because they use a direct mapping from partial observation to the corresponding full observation in one pass, which is not powerful enough to deal with large visual variations between partial and full observations. In this paper, we propose a sequential decoder that \emph{progressively} anticipates the group representation that is expected to contain rich discriminative information as the fully observed activity using $K$ unrolling stages (see Fig.~\ref{fig:architecture3}). This allows us to create a more powerful model for group activity prediction.

Besides, different from individual action prediction methods~\cite{kong2018pami,wang2019progressive}, people's relations formulated as graphs using Eq.~(\ref{eq:normalizeAPP}), are discriminative information for group activity. Moreover, the group activity varies overtime. It is necessary to predict group representations by anticipating relations in the unobserved stage.
As described in Section 3.1, people's relations can be inferred from their action similarity and relative positions. For example, a partial observation of a volleyball activity is given, which contains run-up of ace spikers and waiting gestures of their opponents. Our model is supposed to predict it as ``spiking'' by the cue that the players are moving towards net with their actions.
Therefore, we develop a sequential decoder as a mixture of two graph auto-encoders: an activity auto-encoder $\mathcal{E}_\text{a}\text{-}\mathcal{D}_\text{a}$ for predicting activity representations and a position auto-encoder $\mathcal{E}_\text{p}\text{-}\mathcal{D}_\text{p}$ for predicting positions of multiple people. 
The two auto-encoders are coupled by the shared latent variables $Z_0$ learned from partial observations.

\textbf{Activity auto-encoder $\mathcal{E}_\text{a}\text{-}\mathcal{D}_\text{a}$.}
Using $K$ activity auto-encoders, the proposed sequential decoder progressively anticipates the activity representation by $K$ unrolling stages. Each activity auto-encoder at the $k$-th stage ($k\in\{1,2,\cdots,K\}$) is fed with the output $\hat{X}_{k}$ of the activity auto-encoder at the previous $(k-1)$-th stage. We use the spatiotemporal action features at the last observed frame $t_0$ as the input of the activity auto-encoder at stage $k=1$. We encode the input $\hat{X}_{k}$ of current unrolling stage to a latent variable $Z^{\text{a}}_k$ by 
\begin{equation}
\label{eq:ae}
Z^a_k=\sigma (G_{k}^{\text{p}}\hat{X}_kU_\text{ep})+\sigma(G_{k}^{\text{a}}\hat{X}_kU_\text{ea}),
\end{equation}
and then decodes the activity representation $\hat{X}_{k+1}$:
\begin{equation}
\label{eq:a_d}
\hat{X}_{k+1}=\sigma (G_{k}^{\text{a}}(Z_0+Z^{\text{a}}_{k}))U_\text{da})+\sigma(G_{k}^{\text{p}}(Z_0+Z^{\text{a}}_{k})U_\text{dp}),
\end{equation}
where $U_\text{ep}$, $U_\text{ea}$, $U_\text{dp}$, $U_\text{da}$ are learnable parameters. 
$\hat{X}_{k+1}$ is the anticipated group features at the $k$-th stage, and is served as the input for the activity auto-encoder at the $(k+1)$-th stage. 
The anticipation of ${\hat{X}}_{k+1}$ is conditioned on latent variables $Z_{k}^a$ and $Z_{0}$, in order to both keep track of the short-term information of the previous unrolling stage and use the long-term spatiotemporal information in the partial observations. $G^\text{a}_k$ and $G^\text{p}_k$ are computed by the generated activity features and positions at the $k$-th stage using similar functions as Eq.~(\ref{eq:normalizeAPP}), respectively (replacing time step $t$ by the stage $k$). 

The benefits of the progressive anticipation using $K$ unrolling stages lie in two aspects. 
First, the temporal dependency of activity evolution is naturally built between successive stages. This allows us to naturally anticipate structured group activity representations for prediction purpose. Second, the prediction granularity can be controlled with the number of unrolling stages $K$. The case when $K=1$ is equivalent to the existing one-pass solution used in \cite{kong2018pami,kong2017deep}.

\textbf{Position auto-encoder $\mathcal{E}_\text{p}\text{-}\mathcal{D}_\text{p}$.}
As described in Section 3.1, the interactions between two people also depend on their relative positions. 
Thus, it is necessary to explicitly anticipate the positions of these people in group activity prediction. 

Similar to activity auto-encoder, the proposed sequential decoder also performs $K$ unrolling stages for position prediction for a group of people using $K$ position auto-encoders. Each position auto-encoder at stage $k$ is fed with the output of its previous auto-encoder at stage $k-1$, and outputs the positions of people. Experimental results in Section~\ref{sec:ablation} show that the anticipated future positions of people help improve performance of group activity prediction. 

The position auto-encoder first encodes the positions $\hat{B}_k$ of multiple people to a latent variable $Z^{\text{p}}_k$ at stage $k$ through graph convolution \cite{kipf2016semi}:
\begin{equation}
\label{eq:posEncoder}
Z^{\text{p}}_{k}=\sigma (G_{k}^{\text{p}}\hat{B}_kV_\text{ep})+\sigma(G_{k}^{\text{a}}\hat{B}_kV_\text{ea}),
\end{equation}
and then decodes the positions $\hat{B}_{k+1}$ for the next stage by
\begin{equation}
\label{eq:posdecoder}
\hat{B}_{k+1}=\sigma (G_{k}^{\text{a}}(Z_0+Z^{\text{p}}_{k}))V_\text{da})+\sigma(G_{k}^{\text{p}}(Z_0+Z^{\text{p}}_{k})V_\text{dp}),
\end{equation}
where $V_\text{ep}$, $V_\text{ea}$, $V_\text{dp}$, $V_\text{da}$ are learnable parameters. $G^\text{p}_k$ and $G^\text{a}_k$ are the same graphs used in the activity auto-encoder. The anticipation of $\hat{B}_{k+1}$ is conditioned on latent variables $Z_{k}^\text{p}$ and $Z_{0}$, in order to both keep track of the short-term position information of the previous unrolling stage and use the long-term spatiotemporal information in the partial observations. 

Position prediction is also benefited by sequential prediction via several unrolling stages, since the prediction granularity can be controlled.
Similar to the activity auto-encoder, the position auto-encoder at stage $k=1$ also takes the positions $B_{t_0}$ of people on the last observed frame as input. 
The activity auto-encoder and the position auto-encoder share the same graphs $G^\text{p}_k$ and $G^\text{a}_k$ and are both conditioned on the latent variable $Z_0$ (see Fig.~\ref{fig:architecture3}).

\subsection{Feature Aggregation for Prediction}
SRAM returns both group activity and position representations at each of $K$ unrolling stages. The $K$-th stage corresponds to the full observation status, which contains the most discriminative information of an activity.
We disregard all the outputs given by the activity autoencoders from the $1$-st to $(K-1)$-th stages, and perform max-pooling on the output $\hat{X}_{K+1}$ given by the activity autoencoder at the $K$-th stage as the group activity representations. The resulting feature vector is used for group activity prediction. Similarly, we directly use the output $\hat{B}_{K+1}$ given by the $K$-th position autoencoder to perform position prediction.

\subsection{Loss Functions and Model Learning}

\textbf{Adversarial loss.}
Inspired by \cite{GoodfellowNIPS2014}, we encourage SRAM to generate representations corresponding to ground-truth full observations. 
We use two discriminators for the features generated by the sequential decoder. Discriminator $D_1$ is an activity classifier implemented by a softmax layer. $\mathcal{L}_{\text{cls}}$ is computed on the output of $D_1$. Discriminator $D_2$ is an adversarial regularizer and tells the difference between the generated group features $\hat{X}_{1:K}$ and group features of full videos $F_{1:K}(X)$. Using the adversarial loss, SRAM is encouraged to generate features that are indistinguishable from the group features of the corresponding full videos:
\begin{align}
\mathcal{L}_{\text{GAN}} =& \mathbb{E}_{X_{(1:T)}\sim p_{\text{data}}(X_{(1:T)})}\log 
D_2\left(F_{1:K}(X) \right )\\
+&\mathbb{E}_{X_{(1:t_0)}\sim p_{\text{data}}(X_{(1:t_0)})}\log\big(1-D_{2}(\hat{X}_{1:K}))\nonumber.
\label{eq:advloss}
\end{align}
Note that the generated group representation $\hat{X}_{1:K}$ is computed by SRAM $S$ from the partial observation $X_{1:t_0}$.

\textbf{Sequential reconstruction loss} is proposed to align the predicted activity representations to become close to the ground-truth activity representations at each unrolling stage. Since our method has $K$-stage sequential prediction, it is necessary to encourage the predicted group representations $\hat{X}_{1:K}$ on each of the $K$ unrolling stages to become close to the ground-truth features at that timestamp. 
This is different from adversarial loss that only align the generated features to be close to ground-truth at full observation stage.

We train a separate ST-GCN $F(\cdot)$ as a recognition model to obtain the group activity representations of full videos $X$ for training. The resulting frame-wise group representations $F(X)$ are used to encourage the activity features of the $i$-th person generated at the $k$-th unrolling stage to be similar to the ground-truth features using

\begin{equation}
\label{eq:d2}
\mathcal{L}_{\text{rec}}=\frac{1}{K\times N}\sum_{k=1}^{K}\sum_{i=1}^{N}\left\|\mathbf{\hat{x}}_k(i)-F_k(X,i)\right\|^{2}.
\end{equation}
Here, $F_k(X,i)$ is the features of the $i$-th person of the full video at the $k$-th stage. This loss function sequentially computes the difference between the predicted features $\mathbf{\hat{x}}_k(i)$ (the $i$-th row on $\hat{X}_{1:K}$) for the $i$-th person at unrolling stage $k$ and the ground-truth features $F_k(X,i)$, mimicking how a partial observation is progressively approaching its corresponding full observation.

\textbf{Position regression loss.}
We use the tracklets of individuals provided by~\cite{ibrahim2018hierarchical} as the ground-truth of individuals positions. 
During training, we use the mean square error between the predicted positions and ground-truth positions at the $K$ unrolling stages as loss function:
\begin{equation}
\label{reg_loss}
\mathcal{L}_{\text{reg}}=\frac{1}{K\times N}\sum_{k=1}^{K}\sum_{i=1}^{N}||\hat{\mathbf{b}_k}(i)-\mathbf{b}_k(i)||^2,
\end{equation}
where the predicted position $\hat{\mathbf{b}_k}(i)$ is the $i$-th row of $\hat{B}_k$, i.e., the $i$-th person's position predicted by the sequential decoder at the $k$-th stage.

\textbf{Model learning.}
During training, the overall objective function is written as a sum of
sequential reconstruction loss $\mathcal{L}_{\text{rec}}$,
adversarial loss $\mathcal{L}_{\text{GAN}}$,
classification loss $\mathcal{L}_{\text{cls}}$ implemented by softmax loss, 
and position regression loss $\mathcal{L}_{\text{reg}}$:
\begin{equation}
\min_{\mathcal{E,D}}\;\max_{D_{1},D_{2}}\mathcal{L}_{\text{rec}}+ \mathcal{L}_{\text{GAN}} + \mathcal{L}_{\text{cls}}+\mathcal{L}_{\text{reg}}.
\end{equation}
Sequential relational anticipation model $\mathcal{(E,D)}$ and two discriminators ($D_{1},D_{2}$) are alternatively trained until convergence.

\subsection{Discussion}

\textbf{Group activity modeling and anticipation.} Our SRAM captures the interactions of multiple people in the observation encoder, and anticipates their future relations by a sequential decoder. This is different from existing action prediction methods~\cite{kong2018pami,qi2018stagnet} that can only predict the action of an individual. We believe such a novel method will pave the way for future research in other structured visual prediction.

\textbf{Structured sequential prediction.} Compared with group activity recognition methods~\cite{wu2019learning,qi2018stagnet,ibrahim2018hierarchical}, our method performs \textit{sequential prediction} of group activity, in form of future positions and activity representations. Our activity prediction is also facilitated by explicitly predicting people's future positions.

\textbf{Activity evolution over time.} 
Our sequential decoder \textit{progressively} predicts future representations through several unrolling stages, which boosts performance when only few frames are observed. It is guided by a sequential reconstruction loss, mimicking how a partial observation is sequentially approaching its full observation and an adversarial loss to make the generated full observation features to become indistinguishable from the real full observation features.

\section{Experiments}

\subsection{Datasets}
\textbf{Volleyball Dataset} \cite{ibrahim2016hierarchical} consists of $4830$ video clips distributed in $8$ group activities, such as \emph{left spiking} and \emph{right setting}. Each clip has $41$ frames. \cite{ibrahim2016hierarchical} provides the players' tracklets and splits the dataset into training, validation and testing sets. 
Existing group activity recognition methods~\cite{wu2019learning,ibrahim2016hierarchical,ibrahim2018hierarchical,qi2018stagnet} use the middle $10$ frames of each video. To generalize it to prediction task, we extend it to use the middle $20$ frames as full observations, in order to model sequential dynamics.
Note that the middle $20$ frames contain complete group activity executions, because athletes generally move quickly to complete a group activity, such as direct spiking in a volleyball game.

\textbf{Collective Activity Dataset} (CAD)~\cite{choi2009they} contains $44$ videos with $5$ group activities, including \emph{crossing}, \emph{queueing}, \emph{walking}, \emph{talking} and \emph{waiting}. The group activities in CAD are labeled as the majority of people's individual actions. We use the existing tracklet information and training/testing splits following~\cite{wu2019learning}. The number of the frames in videos ranges from $100$ to $2000$. Following~\cite{qi2018stagnet,wu2019learning,bagautdinov2017social}, we divide each video into $10$-frame video clips. This expands training and testing data to $1746$ and $765$ clips, respectively. CAD mainly contains periodic activities such as \emph{walking}, in which significant changes can be seen in $10$ frames.

\subsection{Implementation Details}
Following~\cite{wu2019learning}, we extract a $1024$-dimensional feature vector for each individual with tracklets provided by~\cite{ibrahim2016hierarchical}, using Inception-v3~\cite{szegedy2016rethinking} as backbone and ROIAlign~\cite{he2017mask}. We use three steps for training: 
First, Inception-v3 pretrained on ImageNet is fine-tuned on single frames by jointly predicting individual actions and group activities. 
Then, we freeze the backbone and finetune the recognition model $F(\cdot$) given full videos in the training set. The recognition model contain two ST-GCN layers~\cite{yan2018spatial}, both with $256$-d hidden units.
After that, we train the proposed model. The observation encoder has two layers ST-GCN with both $256$-d hidden units. The activity auto-encoder's encoder has one graph convolution layer that encodes the input into $256$-d latent feature space. The position auto-encoder has two layer graph convolution by encoding the $2$-d positions into $64$-d space and then $256$-d latent space. During training, SRAM plus classifier $D_1$ and discriminator $D_2$ are alternatively updated.

The experiments are conducted with $10$ different observation ratios ranging from $10\%$ to $100\%$ of full videos length. The number of unrolling stages $K$ is set to $5$. We use stochastic gradient descent for optimization.
For Volleyball dataset, the three steps are trained for $30$ epochs, $10$ epochs and $20$ epochs with learning rate $0.001$, $0.001$, $0.0001$ respectively. 
For Collective Activity Dataset, the three steps are trained for $20$ epochs, $50$ epochs and $10$ epochs with learning rate $0.0001$, $0.0001$, $0.0005$, respectively.

\subsection{Comparison with State-of-the-art}

We compare our method with the state-of-the-art prediction methods LRCN~\cite{tran2015learning}, DeepSCN~\cite{kong2017deep}, IBoW and DBoW~\cite{ryoo2011human}, KD~\cite{wang2019progressive}, AAPNet~\cite{kong2018pami} and state-of-the-art group activity recognition methods, including HRN~\cite{ibrahim2018hierarchical}, HDTM~\cite{ibrahim2016hierarchical} ARG~\cite{wu2019learning}, SSU~\cite{bagautdinov2017social}. 
Following these methods' original setting, LRCN and HDTM adopt the AlexNet~\cite{krizhevsky2012imagenet} as the backbone. HRN, IBow, DBow, DeepSCN and original AAPNet use VGG-19. Our method follows ARG and SSU to use Inception-V3 method. 
HRN, HDTM, ARG, SSU and our method adopt the tracklets of players provided by~\cite{ibrahim2016hierarchical}. To make a fair comparison, we extend state-of-the-art action prediction method AAPNet (``e-AAPNet'' for simplification) to make use of tracking information and use Inception-V3 as backbone.
We train all of the comparison methods using the parameters described in their original papers.

\subsubsection{Results on Volleyball dataset.}
Table~\ref{table:comparison} summarizes the prediction performance of the proposed method, existing action prediction methods and group activity recognition methods.
Results demonstrate that our model outperforms the comparison methods.
Existing action prediction methods, e.g., LRCN, IBoW, DBoW, DeepSCN, AAPNet, KD propose to improve the prediction performance by information transfer.
However, they regard multiple people as a single entity and do not consider the interactions between multiple people.
Thus, the extracted features do not contain informative cues of the interactions of people, resulting in a low prediction performance. 
The proposed method uses tracklets~\cite{ibrahim2016hierarchical}, while the existing predictors for individuals e.g. LRCN, IBoW, DBoW, DeepSCN, AAPNet, KD do not. 
To make a fair comparison, we extend AAPNet to use tracklet information. Experimental results show that our method can predict the dynamics of interactions and better enrich partial observations.

\begin{table*}[t]
\scriptsize
\setlength{\tabcolsep}{1pt}
\caption{Group activity prediction accuracy (\%) on Volleyball dataset with observation ratios ranging from $10\%$ to $100\%$. Group activity recognition results can be seen from the last column, in which $100\%$ frames are observed.}
\begin{tabular}{l|c|c|cccccccccc}
\hline
Models    & Tracklet  & Backbones   & $10\%$   & $20\%$    & $30\%$    & $40\%$     & $50\%$     & $60\%$     & $70\%$      & $80\%$  & $90\%$   & $100\%$ \\ \hline
LRCN~\cite{donahue2015long}   & No	& AlexNet & 48.17	& 51.61	& 54.67	& 57.44	& 59.76	& 61.23 & 63.75	& 64.32	 & 64.77 & 65.37      \\ 
HDTM~\cite{ibrahim2016hierarchical} & Yes & AlexNet  & 52.43	& 59.09	& 66.04	& 76.37	& 80.48	& 81.82   & 84.07	 & 84.47 & 84.60	   & 84.06  \\
IBoW~\cite{ryoo2011stochastic}  & No  & VGG  & 58.03 	& 60.72	& 64.84	& 65.26	& 67.51	& 70.80	& 73.45	& 74.24	& 74.29  & 75.63 \\
DBoW~\cite{ryoo2011stochastic}  & No  & VGG & 58.03	  & 55.56	  & 56.16	  & 58.93	  & 59.90	& 61.97	& 63.79	& 63.06	& 63.88  & 64.78\\ 
DeepSCN~\cite{kong2017deep}  & No  & VGG &59.46	&62.23	&65.52	&70.38	& 72.55	  &77.37  & 79.75    & 80.35   & 80.31  & 80.78\\
HRN~\cite{ibrahim2018hierarchical} & Yes	& VGG & 52.58	& 56.99	& 64.32	& 74.49	& 76.96	& 80.36  & 83.72	  & 84.74 & 84.08 &  85.30 \\ 
KD~\cite{wang2019progressive} & No & VGG & 65.67  & 67.68  & 70.00   & 70.83  & 71.96   & 72.10  & 73.22   &  73.30  &  73.30  &  73.90\\ 
AAPNet~\cite{kong2018pami} & No  & VGG & 59.53	& 65.37	& 68.29	& 72.25	& 75.24	& 77.79	& 79.91	& 80.25	& 80.18  & 80.78\\ 
e-AAPNet~\cite{kong2018pami}  & Yes  & InceptionV3 & 62.98  & 70.31 & 77.64	& 83.55  & 84.91 & 85.86	  & 87.54	& 87.23	& 87.92 & 89.01	  \\
SSU~\cite{bagautdinov2017social}  & Yes & InceptionV3 & 63.20	& 70.65	& 79.66	& 84.07	& 87.13	& 87.65	& 88.30	& 88.18	& 88.41  & 89.01\\ 
ARG~\cite{wu2019learning} & Yes & InceptionV3 & 64.82	& 69.41	& 76.07	& 79.43	& 82.70	& 83.99	& 85.04	& 85.19	& 85.86  & 85.94\\ 
Ours & Yes & InceptionV3  & $\mathbf{77.86}$	& $\mathbf{82.57}$	& $\mathbf{84.97}$	& $\mathbf{87.06}$	& $\mathbf{88.63}$	& $\mathbf{88.93}$	& $\mathbf{89.08}$	& $\mathbf{88.93}$	& $\mathbf{88.48}$	& $\mathbf{91.97}$	 \\ \hline
\end{tabular}
\label{table:comparison}
\end{table*}

Group activity recognition methods such as HDTM, SSU, HRN, ARG do not have capability of gaining extra information from full activity executions. Thus, when the observation ratio is very low ($10\%$ or $20\%$ observations), their performance is much lower than our method. 
Note that ARG applies random sampling strategy by sampling three frames from an entire video as input. In the comparison experiment, this strategy is applied in each of the partial observations as input. 
The proposed method consistently outperforms ARG, as our method captures the temporal dynamics of multiple people in the group, and sequentially generates features close to the corresponding full observations. It improves the representation power of the partial observations, and facilitates group activity prediction.

\setlength\intextsep{0pt}
\begin{wraptable}{r}{0.42\textwidth}
\small
\setlength{\belowcaptionskip}{0.2cm}
\setlength{\tabcolsep}{0.5mm}
\caption{Prediction accuracy (\%) on Collective Activity Dataset.}
\begin{tabular}{l|c|ccc}
\hline
Models        & Tracklet  & $50\%$    & $100\%$    \\ \hline
ARG~\cite{wu2019learning} & Yes  & $88.10$ & $88.37$    \\ 
DeepSCN~\cite{kong2017deep} & No & $81.31$   & $82.22$ \\
AAPNet~\cite{kong2018pami} & No  & $81.57$ & $82.75$    \\ 
e-AAPNet~\cite{kong2018pami}  & Yes & $86.01$ & $86.67$\\
Ours & Yes & $\mathbf{92.55}$ & $\mathbf{92.81}$  \\  \hline
\end{tabular}
\label{table:cad}
\end{wraptable}

\subsubsection{Results on Collective Activity Dataset.}
Comparison results are listed in Table~\ref{table:cad}. Our method outperforms existing methods ARG, DeepSCN, and AAPNet by a large margin. Given tracklets as input, our method is $6.54\%$ higher than e-AAPNet at $50\%$ observation ratio since the people's actions and relations are predicted in our model. Group activities such as \emph{group walking} are cyclic, and thus the prediction performance of our method at $50\%$ observation ratio is close to the one at $100\%$ observation ratio.

\subsection{Ablation Study}\label{sec:ablation}
We perform detailed ablation studies on the Volleyball dataset to evaluate the contributions of the sequential prediction strategy, as well as the loss functions.

\subsubsection{How much does the prediction loss help?}
The impacts of loss functions are analyzed on Volleyball dataset in detail. The evaluation results can also validate the contributions of the proposed sequential prediction strategy.
We compare the following the variants, including: 
(\textbf{1}) without the position regression loss $\mathcal{L}_{\text{reg}}$ defined in Eq.~(\ref{reg_loss}). In this variant, the position auto-encoder for predicting future positions is not used. During sequential prediction, we replace the individuals' positions in the future frames by the ones given by the last observed frame's. The positions are used for computing $G^{\text{p}}$ for each unrolling stage.
(\textbf{2}) without adversarial loss $\mathcal{L}_{\text{GAN}}$.
(\textbf{3}) without sequential reconstruction loss $\mathcal{L}_{\text{rec}}$ for generated features of unrolling stages.
(\textbf{4}) The proposed full network.

Compared with variant (\textbf{1}), the significant performance gains with all different observation ratios show that the prediction of people's positions is of high importance for group activity prediction. Compared with variants (\textbf{2}) and (\textbf{3}), it shows that the adversarial loss $\mathcal{L}_{\text{GAN}}$ and the reconstruction loss $\mathcal{L}_{\text{rec}}$ in our method improve the performance by $0.55\%$ and $0.63\%$ on average, respectively. Therefore, the proposed sequential decoder guided by $\mathcal{L}_{\text{GAN}}$ and $\mathcal{L}_{\text{rec}}$ can generate more discriminative activity representations at each stage.

\subsubsection{How much does the sequential prediction help?}
Our sequential decoder predicts group activity representations at $K$ unrolling stages.
In this experiment, we evaluate the effect of the number of unrolling stages $K$ on the prediction performance.
We set $K$ to $1$, $2$, $5$, and $10$, and compare the prediction performance. Table~\ref{table:stagenumber} indicates that the best overall prediction performance is achieved when $K=5$. The prediction performance is slightly affected when $K=10$, but the computational complexity of the prediction model is increased due to the extra unrolling stages. 
The average prediction performance drops to $81.47\%$ if $K=1$. The variant with $K=1$ is the one that directly maps  partial observation in one unrolling stage, similar to what \cite{kong2018pami,kong2017deep} do.
The result demonstrates the superiority of our progressive prediction in anticipating discriminative group representations given partial observations.
If more stages are allowed ($K=5$ or $K=10$), the sequential decoder in our model can progressively generate discriminative features for group activity prediction even though it is given very limited frames. Therefore, its prediction performance is improved.

\begin{table*}[t]
\centering
\footnotesize
\caption{Ablation studies on volleyball dataset. We show the accuracy($\%$) given videos of observation ratio at $10\%$, $40\%$, $70\%$.}
~\\[6pt]
\setlength{\abovecaptionskip}{0.2cm}
\setlength{\belowcaptionskip}{0.2cm}
\setlength{\tabcolsep}{0.45mm}
\subfloat[\small{Comparison of different loss functions, $\mathcal{L}_{\text{GAN}}$, $\mathcal{L}_{\text{rec}}$, $\mathcal{L}_{\text{reg}}$, and $\mathcal{L}_{\text{cls}}$.}
\label{table:loss}]{
\begin{tabular}[b]{l|cccc}
\hline
 Loss       & $10\%$     & $40\%$       & $70\%$         & Average  \\\hline
($\mathbf{1}$)$\mathcal{L}_{\text{GAN}}$+$\mathcal{L}_{\text{rec}}$+$\mathcal{L}_{\text{cls}}$   & $75.09$  & $85.59$	& $87.06$	   & $84.85$ \\
($\mathbf{2}$) $\mathcal{L}_{\text{rec}}$+$\mathcal{L}_{\text{reg}}$+$\mathcal{L}_{\text{cls}}$   & $76.14$  & $85.79$	& $88.18$	   & $86.30$    \\ 
($\mathbf{3}$) $\mathcal{L}_{\text{reg}}$+$\mathcal{L}_{\text{GAN}}$+$\mathcal{L}_{\text{cls}}$    & $77.61$  & $83.22$	& $85.64$	   & $86.22$ \\
($\mathbf{4}$) Ours        & $\mathbf{77.86}$  & $\mathbf{87.67}$	& $\mathbf{89.08}$	   & $\mathbf{86.85}$	  \\ \hline
\end{tabular}}
\hspace{3mm}
\subfloat[\small{Comparison on the number of unrolling stages $K$.}
\label{table:stagenumber}]{
\begin{tabular}[b]{c|cccc}
\hline
% & & & & \\ 
$K$  & $10\%$ & $40\%$ & $70\%$ & Average \\ \hline
$1$ & $70.38$   &  $80.02$   & $86.14$   &  $81.47$\\ 
$2$  &   $72.36$    &  $86.59$    &  $89.07$     &    $85.22$     \\
$5$  &   $77.86$    &   $\mathbf{87.06}$    &  $89.08$     & $\mathbf{86.85}$  \\
$10$ &   $\mathbf{77.93}$   &   $86.69$    &  $\mathbf{89.23}$     &   $86.79$      \\ \hline
\end{tabular}}
\label{table:ablations}
\end{table*}

\subsection{Position Prediction Evaluation}

\subsubsection{Visualization of predicted positions}

\begin{figure*}[!t]
\begin{center}
%  \fbox{\rule{0pt}{2in} \rule{0.9\linewidth}{0pt}}
\includegraphics[width=1.0\linewidth]{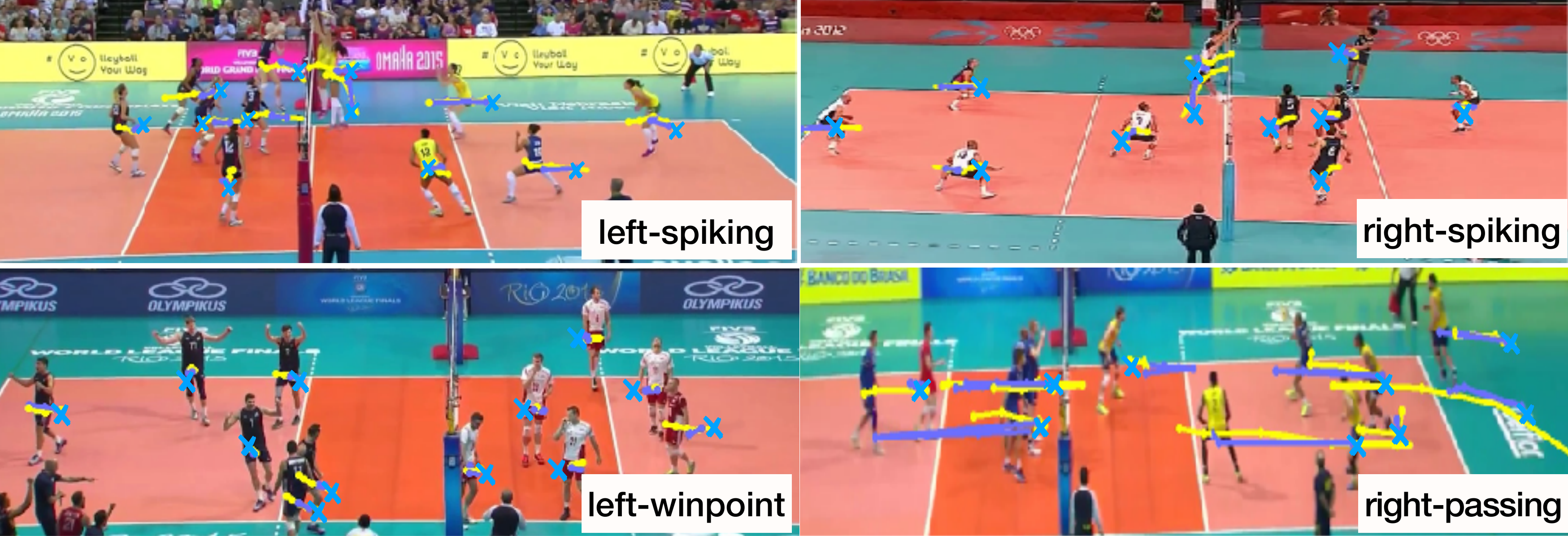}
\end{center}
\caption{Visualization of position predictions. The blue and the yellow lines denote the prediction positions and ground-truth positions~\cite{ibrahim2016hierarchical}, respectively.
``$\times $'' indicates the starting point of the movement. Best viewed in color.
}
\label{fig:visualziation}
\end{figure*}

As shown in Fig.~\ref{fig:visualziation}, we visualize the movement of individuals learned by the position auto-encoder in SRAM. The position auto-encoder progressively predicts the positions of individuals at the unrolling stages. The visualization result shows our position auto-encoder can successfully predict the directions and step sizes of individuals in the future based on partial observations. Although Fig.~\ref{fig:visualziation} (bottom-right) shows the direction of the predicted movement is mostly accurate, the future position of a person is not accurate if the person moves very fast.

\subsubsection{Quantitative evaluation}
We quantitatively evaluate our position prediction results compared to two popular trajectory prediction methods SocialGAN~\cite{gupta2018social} and SocialLSTM~\cite{alahi2016social}. 
Following SocialGAN, Final Displacement Error (FDE) is used to compute the Euclidean distance between the predicted positions and ground-truth positions at the final timestamp and Average Displacement Error (ADE) is used to compute that at each unrolling stage.

\begin{wraptable}{r}{0.42\textwidth}
\setlength{\tabcolsep}{2mm}
\centering
\footnotesize
\caption{\footnotesize{Final Displacement Error (FDE) and Average Displacement Error (ADE) for position prediction.}}
\begin{tabular}{l|l|l}
\hline
Method & FDE & ADE \\ \hline
SocialGAN~\cite{gupta2018social}  &  $5.32$    & $3.05$ \\
SocialLSTM~\cite{alahi2016social} &  $6.44$  & $4.44$\\
Ours   &  $\mathbf{3.62}$ & $\mathbf{2.44}$ \\ \hline
\end{tabular}
\label{table: TRAJ}
\end{wraptable}

As shown in Tab.~\ref{table: TRAJ}, the results demonstrate that our method can accurately predict the future positions for a group of people, and our method outperforms the two trajectory prediction methods.
This is mainly because we capture the relational action dynamics of multiple people while SocialGAN and SocialLSTM do not.

\section{Conclusion}
We have proposed a novel sequential relational anticipation model (SRAM) to predict group activity given only the beginning frames of an activity execution. Our model captures complex relational dynamics of multiple people in the observed frames. It then anticipates the group representations including group activity features and position features. A novel sequential decoder is proposed to progressively anticipates the group representations through several unrolling stages. Extensive results on two datasets demonstrate that our method significantly outperforms the state-of-the-art methods. Results also validate that the progressive anticipation using multiple unrolling stages facilitates group activity prediction. Further experimental results show that the modeling and prediction of people's positions improves our performance on group activity prediction.
~\\[6pt]
\textbf{Acknowledgement}:
We thank Nvidia for the GPU donation. This research is supported in part by ONR Award N00014-18-1-2875.

% \clearpage

% \bibliography{egbib}
\end{document}